\documentclass{article}
\usepackage{spconf,amsmath,graphicx, listings, color}
\usepackage{todonotes}
\usepackage{booktabs}
\usepackage{pgfplots, pgfplotstable}
\usetikzlibrary{matrix}

\title{wav2letter++: The Fastest Open-source Speech Recognition System}
\name{\begin{tabular}{c}Vineel Pratap, Awni Hannun, Qiantong Xu, Jeff Cai, Jacob Kahn, Gabriel Synnaeve, \\
 Vitaliy Liptchinsky, Ronan Collobert\end{tabular}}
\address{Facebook AI Research}
\begin{document}
\maketitle
\begin{abstract}
This paper introduces wav2letter++, the fastest open-source deep learning speech recognition framework. wav2letter++ is written entirely in C++, and uses the ArrayFire tensor library for maximum efficiency. Here we explain the architecture and design of the wav2letter++ system and compare it to other major open-source speech recognition systems. In some cases wav2letter++ is more than 2$\times$ faster than other optimized frameworks for training end-to-end neural networks for speech recognition. We also show that wav2letter++'s training times scale linearly to 64 GPUs, the highest we tested, for models with 100 million parameters. High-performance frameworks enable fast iteration, which is often a crucial factor in successful research and model tuning on new datasets and tasks.
\end{abstract}
\section{Introduction}
\label{sec:intro}
With the growing interest in automatic speech recognition (ASR), the open-source software ecosystem has seen a proliferation of ASR systems and toolkits, including Kaldi~\cite{povey2011kaldi}, ESPNet~\cite{watanabe2018espnet}, OpenSeq2Seq~\cite{openseq2seq} and Eesen\cite{miao2015eesen}. 
Over the last decade these frameworks have shifted from traditional speech recognition based on Hidden Markov Models (HMM) and Gaussian Mixture Models (GMM) to end-to-end neural network based systems. Many of the recent open-source ASR toolkits, including the one presented in this paper, rely on end-to-end acoustic modeling based on graphemes rather than phonemes.
The reason for this shift is two-fold: end-to-end models are significantly simpler and the gap in accuracy to HMM/GMM systems is rapidly closing.
C++ is the 3rd most popular programming language in the world\footnote{https://www.tiobe.com/tiobe-index/}. It allows for complete resource control for high-performance and mission critical systems, and, in addition, static typing helps with large-scale projects by catching any contract mismatches at compile time. Moreover, native libraries can be easily invoked from virtually any programming language. However, adoption of C++ in the machine learning community has been stalled by the absence of well-defined C++ APIs in mainstream frameworks, and C++ was used mostly for performance critical components. As a code base becomes larger, it also becomes cumbersome and error-prone to switch back and forth between a scripting language and C++. Also, provided the adequate libraries, developing in modern C++ is not much slower than in a scripting language. In this paper, we introduce the first open-source speech recognition system written completely in C++. By using modern C++, we do not sacrifice ease of programming yet maintain the ability to write highly efficient and scalable software. In this work, we focus on the technical aspects of ASR systems, such as training and decoding speed, and scalability. 
The rest of this paper is structured as follows. In Section~\ref{sec:design}, we discuss the design of wav2letter++. In Section~\ref{sec:relwork}, we briefly discuss other existing major open-source systems, and benchmark their performance against ours in Section~\ref{sec:experiments}.
\section{Design}
\label{sec:design}
The design of wav2letter++ is motivated by three requirements. First, the toolkit must be able to efficiently train models on datasets containing many thousands of hours of speech. Second, expressing and incorporating new network architectures, loss functions, and other core operations should be simple. And third, the path from model research to deployment should be straightforward, requiring as little new code as possible while maintaining the flexibility needed for research.
\subsection{ArrayFire Tensor Library}
We use ArrayFire~\cite{arrayfire} as our primary library for tensor operations. We chose ArrayFire for several reasons. ArrayFire is a highly optimized tensor library that can execute on multiple back-ends including a CUDA GPU back-end and a CPU back-end. ArrayFire also uses just-in-time code generation to combine series of simple operations into a single kernel call. This results in faster execution for memory bandwidth bound operations and can reduce peak memory use. Another important feature of ArrayFire is the simple interface for constructing and operating on arrays. Compared to other C++ tensor libraries which also support CUDA, the ArrayFire interface is less verbose and relies on fewer C++ idiosyncrasies.
\subsection{Data Preparation and Feature Extraction}
Our feature extraction supports multiple audio file formats (e.g. wav, flac... / mono, stereo / int, float) and several feature types including the raw audio, a linearly scaled power spectrum, log-Mels (MFSC) and MFCCs. We use the FFTW library to compute discrete Fourier transforms~\cite{FFTW05}.
Data loading in wav2letter++ computes features on the fly prior to each network evaluation. This makes exploring alternative features simpler, allows for dynamic data augmentation and makes deploying models much easier since the full end-to-end pipeline can run from a single binary. To make this efficient while training models, we load and decode the audio and compute the features asynchronously and in parallel. For the models and batch sizes we tested, the time spent in data loading is negligible.
\begin{figure}
\begin{center}
  \includegraphics[width=0.72\columnwidth]{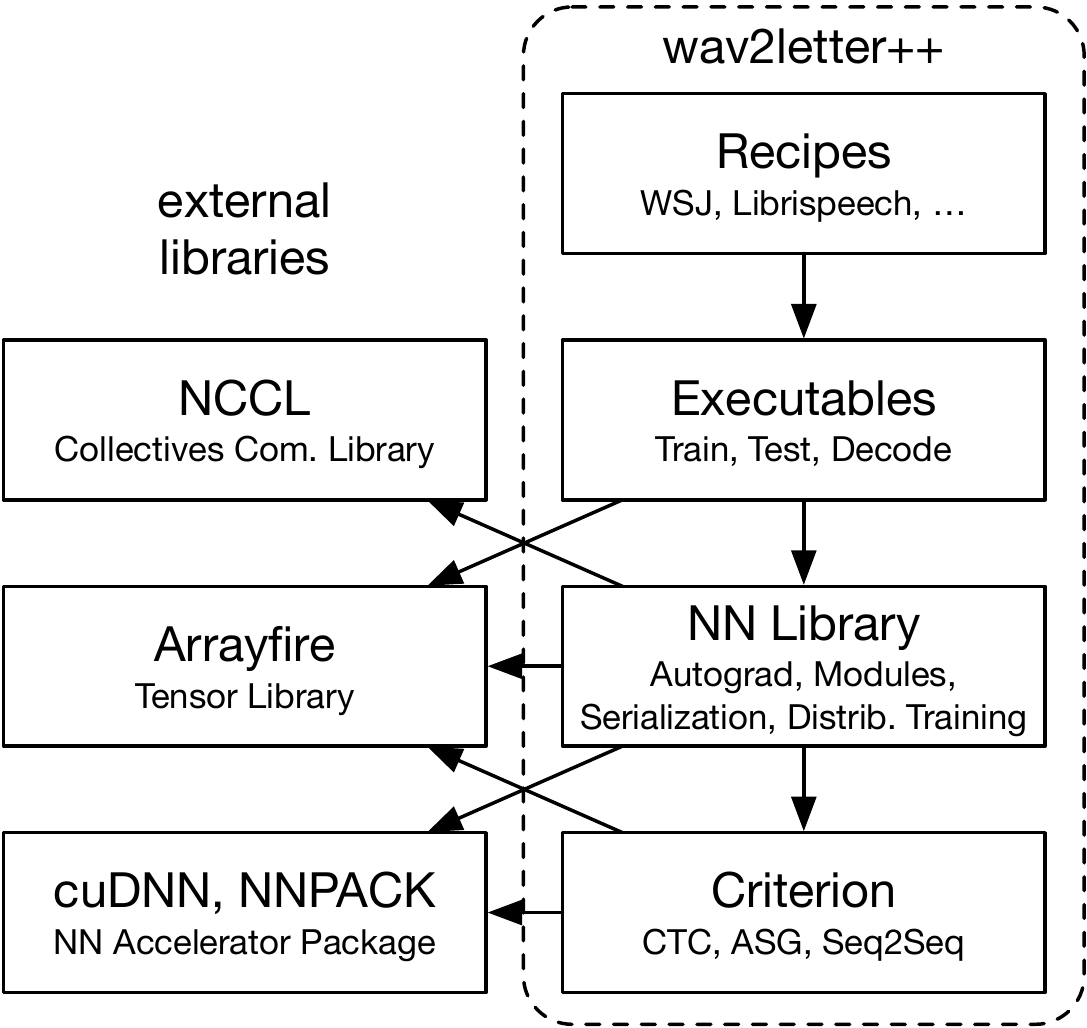}
  \caption{The wav2letter++ library architecture.}
  \label{fig:design}
\end{center}
\end{figure}
 
\subsection{Models}
We support several end-to-end sequence models. Every model is divided into a network and criterion. The network is a function of just the input whereas the criterion is a function of both the input and the target transcription. While the network always has parameters, the parameters of the criterion are optional. This abstraction allows us to easily train different models with the same training pipeline. The supported criteria include Connectionist Temporal Classification (CTC)~\cite{graves2006connectionist}, the original wav2letter AutoSegCriterion (ASG) ~\cite{collobert:2016}, and Sequence-to-Sequence models with attention (S2S)~\cite{bahdanau2014neural, chorowski2015attention}. The CTC criterion does not have parameters whereas the ASG and S2S criteria both have parameters which can be learned. Furthermore, we note that adding new sequence criteria is particularly easy given that loss functions like ASG and CTC can be efficiently implemented in C++.
We support a wide range of network architectures and activation functions -- too many to list here. For certain operations we extend the core ArrayFire CUDA back-end with more efficient cuDNN operations~\cite{chetlur2014cudnn}. We use the 1D and 2D convolutions and the RNN routines provided by cuDNN, among others. Since the network library we use provides dynamic graph construction and automatic differentiation, building new layers or other primitive operations requires little effort.
We give an example showing how to build and train a one layer MLP with the binary cross-entropy loss (in Fig.~\ref{mlp_code}) to demonstrate the simplicity of the C++ interface.
\begin{figure}
\begin{small}
\begin{lstlisting}[language=C++, frame=single,
keywordstyle=\bfseries\color{green!40!black},
commentstyle=\itshape\color{purple!40!black}]
Variable forward(const Variable& x) {
  auto hidden = matmul(weights[0], x);
  hidden = max(hidden, 0); // ReLU
  return matmul(weights[1], hidden);
}
Variable criterion(const Variable& yhat,
                   const Variable& y) {
  auto probs = sigmoid(yhat);
  return -(y * log(probs) +
         (1 - y) * log(1 - probs));
}
for (const auto& xy : trainSet) {
  criterion(forward(xy[0]), xy[1]).backward();
  for (auto& w : weights) {
    w -= lr * w.grad();
    w.zeroGrad();
  }
}
\end{lstlisting}
\end{small}
\vspace{-0.3cm}
\caption{Example: one hidden layer MLP trained with binary cross-entropy and SGD, using automatic differentiation.}
\label{mlp_code}
\end{figure}
\subsection{Training and Scale}
Our training pipeline gives maximum flexibility for the user to experiment with different features, architectures and optimization parameters. Training can be run in three modes - \texttt{train} (flat-start training), \texttt{continue} (continuing with a checkpoint state), and \texttt{fork} (for e.g. transfer learning). We support standard optimization algorithms including SGD and other commonly used first-order gradient-based optimizers.
We scale wav2letter++ to larger datasets with data-parallel, synchronous SGD. For inter-process communication we use the NVIDIA Collective Communication Library (NCCL2)\footnote{https://github.com/NVIDIA/nccl}. To minimize wait time between processes and improve the efficiency of a single process, we sort the dataset on input length prior to constructing batches for training~\cite{hannun2014deep}.
\subsection{Decoding}
\label{sec:decoding}
The wav2letter++ decoder is a beam-search decoder with several optimizations to improve efficiency~\cite{liptchinsky:2017}. We use the same decoding objective as~\cite{liptchinsky:2017}, which includes the constraint from a language model and a word insertion term.
The decoder interface accepts as input the emissions and (if relevant) transitions from the acoustic model. We also give the decoder a Trie which contains the word dictionary and a language model. We support any type of language model which exposes the interface required by our decoder including n-gram LMs and any other stateless parametric LM. We provide a thin wrapper on top of KenLM for n-gram language models~\cite{heafield2011kenlm}.
\begin{table}[]
\begin{small}
\begin{tabular}{lccc}
\toprule
 \textbf{Name} & \textbf{Language} & \textbf{Model(s)} & \textbf{ML Syst.}  \\ \midrule
 Kaldi & C++, Bash & HMM/GMM & - \\
       &           & DNN/LF-MMI      & - \\
 ESPNet & Python, & CTC, seq2seq, & PyTorch,  \\
 & Bash & hybrid & Chainer  \\
 OpenSeq2Seq & Python, C++ & CTC, seq2seq & TensorFlow  \\
 wav2letter++ & C++ & CTC, seq2seq, & ArrayFire \\
  &  & ASG &  \\
  \bottomrule
\end{tabular}
\caption[]{Major open-source speech recognition systems. }
\label{table:taxonomy}
\end{small}
\end{table}
\section{Related Work}
\label{sec:relwork}
We give a brief overview of other commonly used open-source speech recognition systems, including Kaldi~\cite{povey2011kaldi}, ESPNet~\cite{watanabe2018espnet}, and OpenSeq2Seq~\cite{openseq2seq}.
The Kaldi Speech Recognition Toolkit~\cite{povey2011kaldi} is by far the oldest of the aforementioned and consists of a set of stand-alone command-line tools. Kaldi supports HMM/GMM and hybrid HMM/NN-based acoustic modeling, and includes phone-based recipes.
End-to-End Speech Processing Toolkit (ESPNet)~\cite{watanabe2018espnet} tightly integrates with Kaldi and uses it for feature extraction and data pre-processing. ESPNet uses Chainer~\cite{tokui2015chainer} or PyTorch~\cite{pytorch} as a back-end to train acoustic models. It is mostly written in Python, however, following the style of Kaldi, high-level work-flows are expressed in bash scripts. While encouraging the decoupling of system components, this approach lacks the benefit of statically-typed object-oriented programming languages in expressing type-safe, readable and intuitive interfaces. 
ESPNet features both CTC-based~\cite{graves2006connectionist} and attention-based encoder-decoder~\cite{chorowski2015attention} implementations as well as a hybrid model combining both criteria. 
OpenSeq2Seq~\cite{openseq2seq}, similarly to ESPNet, features both CTC-based and encoder-decoder models and is written in Python, using TensorFlow~\cite{abadi2016tensorflow} rather than PyTorch as the back-end. For high-level workflows OpenSeq2Seq also relies on bash scripts that call Perl and Python scripts. A notable feature of the OpenSeq2Seq system is its support for mixed-precision training. Also, both ESPNet and OpenSeq2Seq support models for Text-To-Speech (TTS).
Table~\ref{table:taxonomy} depicts the taxonomy of these open-source speech processing systems. As the table shows, wav2letter++ is the only framework written entirely in C++, which (i) enables easy integration into existing applications implemented virtually in any programming language; (ii) better supports large-scale development with static typing and object oriented programming; (iii) allows for maximum efficiency as discussed in Section~\ref{sec:experiments}. In contrast, dynamically-typed languages such as Python promote quick prototyping, but the lack of enforced static typing often hinders large-scale development.
\section{Experiments}
\label{sec:experiments}
In this section we discuss the performance of ESPNet, Kaldi, OpenSeq2Seq and wav2letter++ in a comparative study. The ASR systems are evaluated on the large vocabulary task of the Wall Street Journal (WSJ) dataset~\cite{wsj}. We measure both the average epoch time on WSJ during training and the average utterance decoding latency. 
The machines we use for the experiments have the following hardware configuration: each machine features eight NVIDIA Tesla V100 Tensor Core GPUs on NVIDIA SXM2 Modules with 16GB of memory. Each compute node has 2 Intel Xeon E5-2698 v4 CPUs, totalling 40 (2$\times$20) cores, 80 hardware threads (``cores''), at 2.20GHz. All machines are connected over a 100Gbps InfiniBand network.
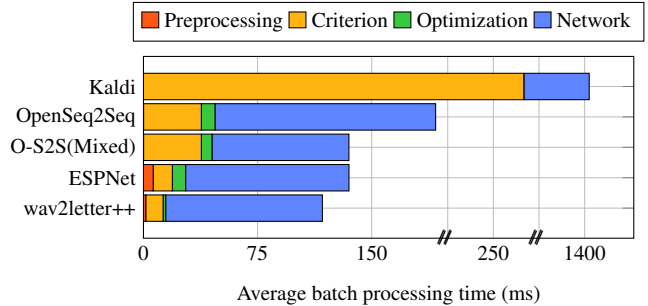
\begin{figure}
\begin{footnotesize}
\definecolor{featurization}{RGB}{255,87,18}
\definecolor{criterion}{RGB}{255,180,18}
\definecolor{optimization}{RGB}{56,201,65}
\definecolor{network}{RGB}{95,132,255}

\begin{tikzpicture}
\begin{axis}[
    xbar stacked,   
    xmin=0,         
    ytick=data,     
    width=0.72\columnwidth,
    yticklabels from table={timings.dat}{System},  
    xlabel=Average batch processing time (ms),
    grid=major,
    height=4.cm,
    width=8.1cm,
    enlarge y limits={abs=0.4cm},
    legend style={
        legend columns=-1,
        at={
            (0.5,1.3),
        },
        anchor=north
    },
    cycle list={
        {fill=featurization},
        {fill=criterion},
        {fill=optimization},
        {fill=network},
    },
    xticklabels={0,75,150,,250,,1400},
    xtick={0,75,150,
        200, 
        230, 
        260, 
        290  
    },
    xtick style={draw=none},
    try min ticks=9,
    x tick label style={rotate=0},
    extra x ticks={198,202,258,262},
    extra x tick style={grid=none, draw=none, tick label style={xshift=-0.04cm,yshift=.21cm, rotate=175}},
    extra x tick label={\color{black}{/\!\!/}}
]
\addplot table[x=Featurization, y expr=\coordindex]{timings.dat};
\addplot table[x=Criterion, y expr=\coordindex]{timings.dat};
\addplot table[x=Optimization, y expr=\coordindex]{timings.dat};
\addplot table[x=Network, y expr=\coordindex]{timings.dat};
\legend{Preprocessing,Criterion,Optimization,Network};
\end{axis}
\end{tikzpicture}
\vspace{-0.5cm}
\end{footnotesize}
  \caption{Time in milliseconds for the major steps in the training loop. The times are averaged for each batch over a full epoch.}
\label{fig:breakdown}
\end{figure}
 
\subsection{Training}
\label{subsec:resultstraining}
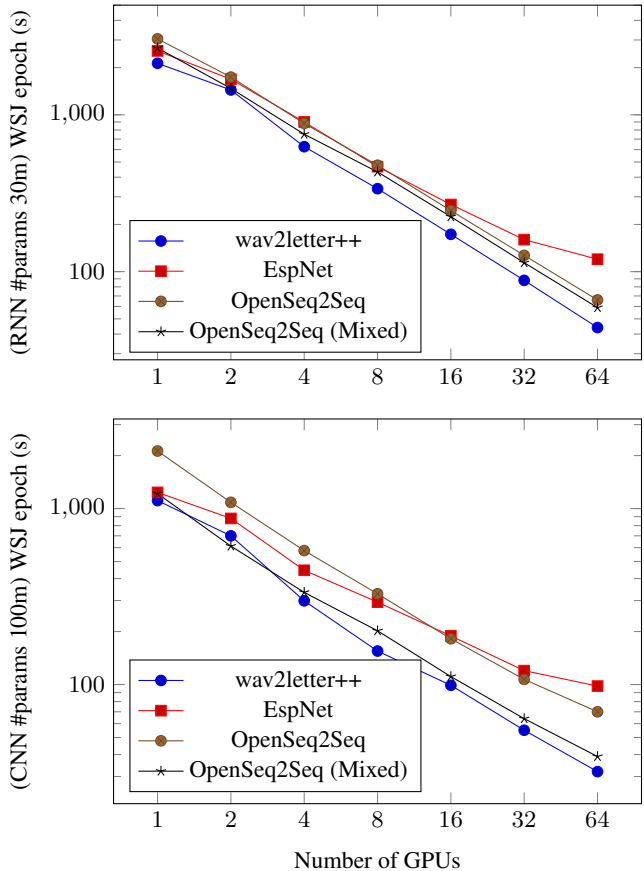
\begin{figure}
\newif\ifgpuxlabel
\gpuxlabelfalse
\begin{small}
\begin{tikzpicture}
\begin{axis}[
  xlabel=\ifgpuxlabel Number of GPUs \else \fi,
  ylabel=(RNN \#params 30m) WSJ epoch (s),
  ymax=5001,
  xtick=data,
  xmode=log,
  log ticks with fixed point,
  ymode=log,
  legend style={at={(0.03,0.02)},anchor=south west},
  width=\columnwidth,
  height=\ifgpuxlabel 6.7cm \else 6.3cm \fi
]
\addplot table [y=wav2letter, x=GPUs]{speed30m.dat};
\addlegendentry{wav2letter++}
\addplot table [y=EspNet, x=GPUs]{speed30m.dat};
\addlegendentry{EspNet}
\addplot table [y=OpenSeq2seq, x=GPUs]{speed30m.dat};
\addlegendentry{OpenSeq2Seq}
\addplot table [y=OpenSeq2seq_mixed, x=GPUs]{speed30m.dat};
\addlegendentry{OpenSeq2Seq (Mixed)}
\end{axis}
\end{tikzpicture}
\gpuxlabeltrue
\begin{tikzpicture}
\begin{axis}[
  xlabel=\ifgpuxlabel Number of GPUs \else \fi,
  ylabel=(CNN \#params 100m) WSJ epoch (s),
  xtick=data,
  xmode=log,
  log ticks with fixed point,
  ymode=log,
  legend style={at={(0.03,0.03)},anchor=south west},
  width=\columnwidth,
  height=\ifgpuxlabel 6.7cm \else 6.3cm \fi
]
\addplot table [y=wav2letter, x=GPUs]{speed100m.dat};
\addlegendentry{wav2letter++}
\addplot table [y=EspNet, x=GPUs]{speed100m.dat};
\addlegendentry{EspNet}
\addplot table [y=OpenSeq2seq, x=GPUs]{speed100m.dat};
\addlegendentry{OpenSeq2Seq}
\addplot table [y=OpenSeq2seq_mixed, x=GPUs]{speed100m.dat};
\addlegendentry{OpenSeq2Seq (Mixed)}
\vspace{-0.5cm}
\end{axis}
\end{tikzpicture}
\end{small}
\vspace{-0.5cm}
  \caption{Comparison of training times (log scale). \textbf{Top}: An RNN with 30m parameters, inspired by DeepSpeech 2~\cite{hannun2014deep}: 2~spatial convolution layers, followed by 5 bidirectional LSTM layers, followed by 2 linear layers. \textbf{Bottom}: A CNN with 100m parameters, similar to~\cite{liptchinsky:2017}: 18 temporal convolution layers, followed by 1 linear layer.}
\label{fig:resultstraining}
\end{figure}

We evaluate training time with respect to both scaling network parameters and increasing the number of GPUs used. We consider 2 types of neural network architectures: recurrent, with 30 million parameters, and purely convolutional, with 100 million parameters, as depicted in the top and bottom charts of Figure~\ref{fig:resultstraining}, respectively. For OpenSeq2Seq we consider both \texttt{float32} as well as mixed precision \texttt{float16} training.  
For both networks, we use 40-dimensional log-mel filterbanks as inputs, and CTC~\cite{graves2006connectionist} as the criterion (CPU-based implementation). For Kaldi, we use the LF-MMI~\cite{povey2016LFMMI} criterion as CTC training is not available in the standard Kaldi recipes. All models are trained with SGD with momentum. We use a batch size of 4 utterances per GPU. Every run is restricted to using 5 CPU cores for each GPU. Figure~\ref{fig:breakdown} provides more detailed look into major components of the training pipeline; the processing time is averaged over entire epoch using a single GPU. 
For both models, wav2letter++ has a clear advantage that increases as we scale out the computation. For smaller models with 30 million parameters, wav2letter++ is more than 15\% faster than the next-best system, even on a single GPU. Note that since we use 8 GPU machines, experiments on 16, 32 and 64 GPUs involve multi-node communication. ESPNet did not support multi-node training out-of-the-box. We extend it by using the PyTorch \texttt{DistributedDataParallel} module with the NCCL2 back-end. ESPNet relies on pre-computed input features, while wav2letter++ and OpenSeq2Seq compute features on the fly for the sake of flexibility. In some cases, mixed precision training decreases the epoch time by more than 1.5x for OpenSeq2Seq. This is an optimization which wav2letter++ can benefit from in the future. The Kaldi recipe for LF-MMI does not synchronize gradients for each SGD update; the per-epoch time is still more than 20x slower. We did not include Kaldi in Figure~\ref{fig:resultstraining} as the criterion (LF-MMI) and optimization algorithm are not easily comparable.
\subsection{Decoding}
\label{subsec:resultsdecoding}
\setlength\tabcolsep{3.8pt}
\begin{table}[]
\begin{small}
\begin{tabular}{lccc}
\toprule
 \textbf{Name} & \textbf{WER (\%)} & \textbf{Time/sample (ms)} & \textbf{Memory (\uppercase{GB})}  \\ \midrule
 ESPNet & 7.20 & 1548 & --\\
 \midrule
 OpenSeq2Seq & 5.00 & 1700 & 7.8 \\
 OpenSeq2Seq & 4.92 & 9500 & 26.6 \\
 \midrule
 wav2letter++ & 5.00 & 10 & 3.9 \\
 wav2letter++ & 4.91 & 140 & 5.5 \\
  \bottomrule
\end{tabular}
\caption[]{Decoding performance on LibriSpeech \emph{dev-clean}. }
\label{table:decodingperf}
\end{small}
\end{table}
wav2letter++ includes a one-pass beam-search decoder (see Section~\ref{sec:decoding}), written in C++. We benchmark it against other beam-search decoders available in OpenSeq2Seq and ESPNet. Kaldi is not included, as it does not support CTC decoding, and implements a WFST-based decoder. We feed each decoder identical, pre-computed emissions generated by a fully-convolutional OpenSeq2Seq model Wave2Letter+\footnote{https://nvidia.github.io/OpenSeq2Seq/html/speech-recognition.html} trained on LibriSpeech. This enables independent measurement of performance given the same model. The 4-gram LibriSpeech language model is used for OpenSeq2Seq and wav2letter++, as ESPNet does not support n-gram LM decoding. In Table~\ref{table:decodingperf}, we report decoding time and peak memory usage, for single thread decoding, on LibriSpeech \emph{dev-clean} to reach a WER of $5.0\%$, and the best-obtainable WER for each framework. Hyper-parameters were heavily tuned such that reported results reflect the best-possible speed for the reported WER. wav2letter++ not only outperforms similar decoders by more than an order of magnitude, but also uses considerably less memory.
\section{Conclusion}
\label{sec:conclusion}
In this paper we introduced wav2letter++: a fast and simple system for developing end-to-end speech recognizers. The framework is written entirely in C++ which makes it efficient to train models and perform real-time decoding. Our initial implementation shows promising results compared to the other speech frameworks; though wav2letter++ can continue to benefit from further optimization. Because of its simple and extensible interface, wav2letter++ is well suited as a platform for rapid research in end-to-end speech recognition. At the same time, we leave open the possibility that certain optimizations might be possible with Python-based ASR systems to narrow the gap with wav2letter++.
\vfill\pagebreak
\bibliographystyle{IEEEbib}
\bibliography{refs}
\end{document}